\RequirePackage{amsmath}
\documentclass[dvipsnames]{svproc}
\PassOptionsToPackage{linktocpage}{hyperref}

\usepackage{graphics}
\usepackage{float}
\usepackage{multirow}
\usepackage{booktabs}
\usepackage[numbers,sectionbib]{natbib}
\usepackage{bibentry}

\usepackage{mathtools}
\usepackage{wrapfig}
\usepackage[dvips]{graphicx}
\usepackage{floatflt,epsfig, wrapfig}
\usepackage{array}
\usepackage{dirtytalk}
\usepackage{ulem}
\usepackage[export]{adjustbox} 
\usepackage{breakcites}
\usepackage{amsmath}
\usepackage{amsfonts}
\usepackage{amssymb}
\usepackage{bbm}
\usepackage[colorinlistoftodos]{todonotes}
\usepackage[colorlinks=true, allcolors=blue]{hyperref}
\usepackage{lscape}
\usepackage{pgfgantt}
\usepackage{esvect}

\usepackage[caption=false]{subfig}

\def\rot#1{\rotatebox{90}{#1}}

\begin{document}
\mainmatter              
\title{Graph-based Topic Extraction from Vector Embeddings of Text Documents: \\ Application to a Corpus of News Articles} 
%
\titlerunning{Graph-based Topic Extraction from Document Vector Embeddings}  
%
\author{M. Tarik Altuncu$^{*}$ \and Sophia N. Yaliraki$^\dagger$ \and Mauricio Barahona$^{*}$}
\authorrunning{Altuncu et al.} 
%
%
\institute{Department of Mathematics$^{*}$ and Department of Chemistry$^{\dagger}$, \\ Imperial College London, London, UK}

\maketitle              

\begin{abstract}
Production of news content is growing at an astonishing rate. 
To help manage and monitor the sheer amount of text, there is an increasing need to develop efficient methods that can provide insights into emerging content areas, and  stratify unstructured corpora of text into `topics' that stem intrinsically from content similarity. 
Here we present an unsupervised framework that brings together powerful vector embeddings from natural language processing  with tools from multiscale graph partitioning that can reveal natural partitions at different resolutions without making \textit{a priori} assumptions about the number of clusters in the corpus.
We show the advantages of graph-based clustering through end-to-end comparisons with other popular clustering and topic modelling methods, and also evaluate different text vector embeddings, from classic Bag-of-Words to Doc2Vec to the recent transformers based model Bert. 
This comparative work is showcased through an analysis of a corpus of US news coverage
during the presidential election year of 2016.
\end{abstract}

\section{Introduction}
\vspace{-.2cm}

The explosion in the amount of news and journalistic content generated across the globe, coupled with extended and instantaneous access to information via online media, makes it difficult and time-consuming to monitor news and opinion formation in real time. 
There is an increasing need for tools that can pre-process, analyse and classify raw text to extract interpretable content; specifically, identifying topics and content-driven groupings of articles. 
This is in contrast with traditional approaches to \textit{text classification}, typically reliant on human-annotated labels of pre-designed categories and hierarchies~\citep{burkhardt_survey_2019}.
Methodologies that provide automatic, unsupervised clustering of articles based on content directly from free text without external labels or categories could thus provide alternative ways to monitor the generation and emergence of news content.

In recent years, fuelled by advances in statistical learning, there has been a surge of research on unsupervised topic extraction from text corpora~\citep{dieng_topic_2020, lenz_measuring_2020}. In previous work~\citep{altuncu_free_2019}, we developed a graph-based framework for the clustering of text documents that combines the advantages of paragraph vector representation of text through Doc2vec~\citep{le_distributed_2014} with Markov Stability (MS) community detection~\citep{delvenne_stability_2010}, a multiscale graph partitioning method that is applied to a geometric similarity graph of documents derived from the high-dimensional vectors representing the text. 
Through this approach, we obtained robust clusters at different resolutions (from finer to coarser) corresponding to groupings of documents with similar content at different granularity (from more specific to more generic). Because the similarity graph of documents is based on vector embeddings that capture syntactic and semantic features of the text, the graph partitions produce clusters of documents with consistent content in an unsupervised manner, i.e., without training a classifier on hand-coded, pre-designed categories.
Our original application in Ref.~\citep{altuncu_free_2019} was a very large and highly specialised corpus of incident reports in a healthcare setting with specific challenges, i.e., highly specialised terminology mixed with informal and irregular use of language and abbreviations.   
The large amount of data (13 million records) in that corpus allowed us to train a specific Doc2vec model to circumvent successfully these restrictions.

In this work, we present the extension of our work in Ref.~\citep{altuncu_free_2019} to a more general use-case setting by focussing on topic extraction from a relatively small number of documents in standard English. We illustrate our approach through a corpus of $\sim$9,000 news articles published in the US during 2016. Although for such small corpora, training a specific language model is less feasible, the use of standard language allows us to employ English language models trained on generic corpora (e.g., Wikipedia articles).
Here, we use pre-trained language models obtained with Doc2Vec~\cite{le_distributed_2014} and with recent transformer models like Bert~\citep{devlin_bert:_2019}, which are based on deeper neural networks, and compare them to classic Bag-of-Words (BoW) based features. 
In addition, we evaluate our multiscale graph-based partitioning (MS) against classic probabilistic topic clustering methods like LDA~\citep{blei_latent_2003} and other widely-used clustering methods (k-means and hierarchical clustering).

    \section{Feature Vector Generation from Free Text}
    \label{sec:news:features}

\vspace{-.25cm}
\paragraph{Data:} 
The corpus consists of 8,748 online news articles published by US-based Vox Media during 2016, a presidential election year in the USA\footnote{The news corpus is accessible on \url{https://data.world/elenadata/vox-articles}}. The corpus excludes articles without specific news content, without an identifiable publication date, or with very short text (less than 30 tokens).

\vspace{-.25cm}
\paragraph{Pre-processing, tokenisation and normalisation:}
We removed HTML tags, code pieces, and repeated wrapper sentences (header or footer scripts, legal notes, directions to interact with multimedia content, signatures).
We replaced accented characters with their nearest ASCII characters; removed white space characters; and divided the corpus into lists of word tokens via regex word token divider `\textbackslash w+'.
To reduce semantic sparsity, we used Part of Speech (POS) tags to leave out all sentence structures but adjectives, nouns, verbs, and adverbs. 
The remaining tokens are lowered and converted to lemmas using the WordNet Lemmatizer~\citep{miller_wordnet:_1995}.
Lastly, we removed common (and thus less meaningful) tokens\footnote{The full list of common words is: \{`be', `have', `do', `make', `get', `more', `even', `also', `just', `much', `other', `n't', `not', `say', `tell', `re'\}}.

\vspace{-.25cm}    
\paragraph{Bag-of-Words based features:}
\label{sec:news:tfidf}
We produce TF-IDF (Term Frequency-Inverse Document Frequency) BoW features filtered with Latent Semantic Analysis (LSA)~\citep{papadimitriou_latent_1998} to reduce the high-dimensional sparse TF-IDF vectors to a 300 dimensional continuous space.

\vspace{-.25cm}   
\paragraph{Neural network based features:}        
Doc2vec~\citep{le_distributed_2014}, a version of Word2vec~\citep{mikolov_distributed_2013} adapted to longer forms of text, was one of the original neural network based methods with the capability to represent full length free text documents into an embedded fixed-dimensional vector space.
Before it can be used to infer a vector for a given document, Doc2Vec must be trained on a corpus of text documents that share similar context and vocabulary with the documents to be embedded.
As a source of common English compatible with the broad usage in news articles,
we used Gensim version 3.8.0~\citep{rehurek_software_2010} to train a Doc2Vec language model on a recent Wikipedia dump consisting of 5.4 million articles pre-processed as described above
\footnote{English Wikipedia corpus (1 December 2019) downloaded from \url{https://dumps.wikimedia.org/enwiki/}.}.
The optimised model had hyperparameters \{training method = dbow, number of dimensions for feature vectors size = 300, number of epochs = 10, window size = 5, minimum count = 20, number of negative samples = 5, random down-sampling threshold for frequent words = 0.001\}.

\vspace{-.25cm}
\paragraph{Transformers based features:}        
Natural Language Processing (NLP) is currently evolving rapidly with the emergence of transformers-based, deep learning methods such as ELMo~\citep{peters_deep_2018}, GPT~\citep{radford_language_2018} and BERT, the recent state-of-the-art model by Google~\citep{devlin_bert:_2019}.
The first step for these models, called pre-training, uses different learning tasks (e.g., next sentence prediction or masked language model) to model the language of the training corpus without explicit supervision. 
Whereas the neural network in Doc2vec only has two layers, transformers-based models involve deeper neural networks, and thus require much more data and massive computational resources to pre-train.
Fortunately, these methods have publically available models for popular languages. Here, we use the model `BERT base, Uncased’ with 12-layers and 100 million parameters, which produces 768-dimensional vectors based on a pre-training using BookCorpus~\citep{zhu_aligning_2015} and English Wikipedia articles.
Because Bert models carry out their own pre-processing (WordPiece tokenisation and Out-Of-Vocabulary handling steps~\citep{schuster_japanese_2012}), we do not apply our tokenisation and normalisation routine.
Since transformers based models cannot process text longer than a few sentences (510 tokens in total) due to memory limits in GPUs, we analyse the sentences in an article individually; obtain embedded vectors for all; and compute the feature vector of the document as the average of sentence vectors. 
We obtained two feature vectors from Bert: (i) {\it baas}: reduced mean vector of each token's embedded vector on the second from last layer among the 12 layers, as given by bert-as-service\footnote{bert-as-service is an open-source library published at \url{github.com/hanxiao/bert-as-service}.};
(ii) {\it sbert}: sentence level vector computed with Sentence Transformers\footnote{Sentence Transformers is an open-source library published at \url{github.com/UKPLab/sentence-transformers}.}~\citep{reimers_sentence-bert:_2019}, which
improves the pre-trained Bert model using a 3-way softmax classifier for Natural Language Inference (NLI) tasks.

Although BERT models are powerful, they are optimised for supervised downstream tasks and need to be fine-tuned further through a secondary step on specifically labelled annotated data sets. Hence pre-trained BERT models without fine-tuning are not optimised for the quality of their vector embedding, which is the primary input for our unsupervised clustering task.

    \section{Finding Topic Clusters Using Feature Vectors}
    \label{sec:news:label}

\vspace{-.2cm}
The above methods lead to five different feature vectors for the documents in the corpus. These vectors are then clustered through the graph-based MS framework, which we benchmark against alternative graph-less clustering methods.

\vspace{-.25cm}   
        \subsection{Graph Construction and Community Detection}

In many real-world applications, there is an absence of accurate prior knowledge or ground truth labels about the underlying data. In such cases, unsupervised clustering methods must be applied. Rather than fixing the number of clusters \textit{a priori}, we use here Markov Stability, a multiscale method that produces a series of intrinsically robust partitions at different levels of resolution. MS thus allows flexibility in choosing the granularity of topics  as appropriate for the analysis.

\vspace{-.25cm}            
\paragraph{Constructing a Sparsified Similarity Graph:}
From the feature vectors of the 8,748 articles, we compute all pairwise cosine similarities. This dense similarity matrix is sparsified by using the MST-kNN method~\citep{veenstra_spectral_2016} to construct a sparse geometric graph.
MST-kNN consists of two steps: obtain the Minimum Spanning Tree (MST) to ensure global connectivity in the graph followed by linking the k Nearest Neighbours (kNN) to each document. so as to connect highly similar documents thus preserving the local geometry. Based on our work in~\citep{altuncu_free_2019}, we set $k=13$.
We construct MST-kNN similarity graphs from each of the five sets of feature vectors: $\mathcal{G}_\textrm{\small \it tfidf}$, $\mathcal{G}_\textrm{\small \it tfidf+lsa}$, $\mathcal{G}_\textrm{\small \it d2v}$, $\mathcal{G}_\textrm{\small \it baas}$, $\mathcal{G}_\textrm{\small \it sbert}$.

\vspace{-.25cm}    
\paragraph{Multiscale Graph Partitioning with MS:}
            
\begin{figure}[h]
    \centering
\includegraphics[trim={10.5cm 1.75cm 10.5cm 2.2cm}, clip, width=.95\linewidth]{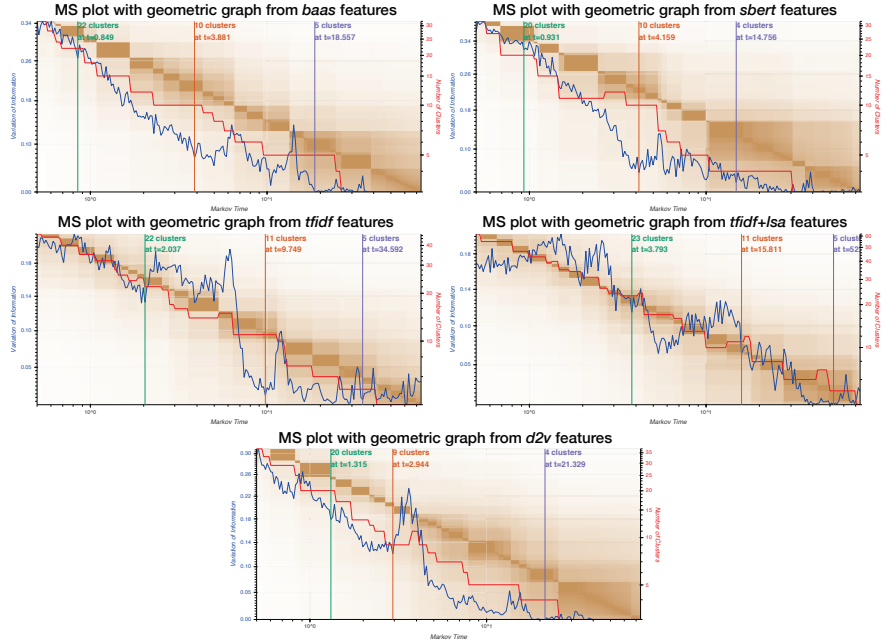}
\caption[Markov Stability plots with selected scales]{
Results of Markov Stability for the five similarity graphs constructed from different document feature vectors.  As a function of Markov time $t$, we show: the number of clusters of the optimised partition $\mathcal{P}_t$ (red line); 
the variation of information $VI(\mathcal{P}_t)$ for the ensemble of optimised solutions at each $t$ (blue line); and the variation of information $VI(\mathcal{P}_t,\mathcal{P}_{t'})$ between the optimised partitions across Markov time (background colourmap). 
Robust partitions correspond to dips of $VI(\mathcal{P}_t)$ and extended plateaux of the number of communities and $VI(\mathcal{P}_t,\mathcal{P}_{t'})$.
The selected partitions are indicated by vertical lines: green (Fine), orange (Medium) and purple (Coarse).
}
\label{fig:news:MS}
\end{figure}
 
We applied the MS partitioning algorithm~\citep{delvenne_stability_2010, lambiotte_random_2014, schaub_markov_2012} to these five similarity graphs. 
We compute the partitions $\mathcal{P}_t$ that optimise the MS cost function as we vary the Markov time $t$, a parameter that modifies the coarseness of the partitions. We choose three partitions at different resolutions that are most robust, both across $t$ and to the optimisation process~\cite{lambiotte_random_2014, schaub_markov_2012}, and label them as fine (F), medium (M), and coarse (C). The three partitions for the five similarity graphs shown in Figure~\ref{fig:news:MS} are:
\begin{enumerate}
    \item $\mathcal{G}_\textrm{\small \it baas}$: $[22,10,5]$ communities at \footnotesize{$t=[0.849,3.881,18.557]$}
    \item $\mathcal{G}_\textrm{\small \it sbert}$: $[20,10,4]$ communities at \footnotesize{$t=[0.931,4.159,14.756]$}
    \item $\mathcal{G}_\textrm{\small \it tfidf}$: $[22,11,5]$ communities at \footnotesize{$t=[2.037,9.749,34.592]$}
    \item $\mathcal{G}_\textrm{\small \it tfidf+lsa}$: $[23,11,5]$ communities at \footnotesize{$t=[3.793,15.811,52.356]$}
    \item $\mathcal{G}_\textrm{\small \it d2v}$: $[20,9,4]$ communities at \footnotesize{$t=[1.315,2.944,21.329]$}.
\end{enumerate}

\vspace{-.25cm}    
        \subsection{Graph-less clustering methods as benchmarks}

We benchmark our graph-based clustering against: (i) two widely used graph-less clustering methods: k-means and hierarchical clustering with Ward linkage. Both methods are applied using the implementation and default parameters in Scikit-Learn version 0.22;
(ii) LDA probabilistic topic models~\citep{blei_latent_2003} for each resolution (F, M, C) using the state-of-the-art implementation, i.e., default parameters of LdaMulticore in Gensim version 3.8.0 setting the number of passes to 100 to enaure convergence.
Unlike the other clustering methods, LDA does not produce a hard assignment of news articles into clusters, but returns probabilities over all clusters. We assign each article to its most probable topic (i.e, cluster).
Because LDA works on term frequencies (\textit{tf}), we cannot use the other feature vectors.

We remark that,  had we not already obtained our MS results, it would not be easy to make an informed  \textit{a priori} choice of number of topics/clusters for any of these methods.
Hence we use the three MS levels (F=20, M=9, C=4) as the number of clusters in the benchmark methods.
Counting five different feature vectors and three clustering algorithms along with the LDA, we have 16 experiments per each resolution level (F,M,C). The next sections are devoted to the quantitative and qualitative comparison of all 48 experiments.

    \section{Evaluation of Topic Clusters}
    \label{sec:news:evaluation}

\vspace{-.25cm}    
    	\subsection{Quantitative Analysis of Partition Quality}
    	\label{sec:news:quantitative}

In the absence of 'ground truth' topics in our corpus, we cannot score directly the quality of the clusters.
Instead, we compute two different measures of the consistency and relevance of cluster content.

\vspace{-.25cm}    
		    \paragraph{Measuring intrinsic topic coherence with the Aggregate PMI: }
		    \label{sec:news:pmi}

Following~\cite{newman_external_2009, newman_automatic_2010} we create an intrinsic measure of topic coherence without reference to external ground truth topics. OUr measure considers the association of frequent `word couples' in each cluster and computes the pointwise mutual information (PMI) compared to its reference score in a standard corpus (in our case, the English Wikipedia).  
The topical coherence of articles within topic averaged over all topics gives the Aggregate PMI scores ($\widehat{\text{PMI}}$) for each experiment is shown in Table~\ref{table:news:PMI_scores_10}.

Our results of topic coherence show that LDA perform poorly as the least coherent topics, whereas MS partitions are the most coherent overall for most types of features (except for a few cases where k-means and hierarchical Ward are better). 
On average across all features,  MS is the best, followed by k-means and Ward.
Regarding text features, there is not a best performing one across all resolutions. Interestingly, 
\textit{tfidf} features produce coherent topic clusters, perhaps due to the fact that
\textit{tfidf} and \textit{tfidf+lsa} features are based on counts of word occurrence, and hence very similar in spirit to PMI scores.
Among the neural network based features, \textit{baas} vectors from Bert work best in the fine and medium levels while \textit{d2v} is best at the coarse level.

\begin{table}[h!]
\centering
\scalebox{0.8}{
\begin{tabular}{ll  l|cccc|cccc|cccc}
\toprule
& &\textit{Resolution: } & \multicolumn{4}{c}{Fine} & \multicolumn{4}{c}{Medium} & \multicolumn{4}{c}{Coarse} \\ 
& &\textit{Clustering: } &      MS &   Ward & k-means &    LDA &      MS &   Ward & k-means &    LDA &      MS &   Ward & k-means &    LDA \\
\midrule
&&baas      &   1.866 &  1.692 &   1.832 &    - &   1.775 &  1.553 &   1.717 &    - &   1.656 &  1.332 &   1.498 &    - \\
&&d2v       &   1.843 &  1.660 &   1.856 &    - &   1.584 &  1.672 &   1.673 &    - &   1.724 &  1.532 &   1.349 &    - \\
&&sbert     &   1.731 &  1.706 &   1.829 &    - &   1.772 &  1.639 &   1.638 &    - &   1.482 &  1.405 &   1.635 &    - \\
&&tf        &     - &    - &     - &  1.473 &     - &    - &     - &  1.292 &     - &    - &     - &  1.182 \\
 \rot{\rlap{~\textit{Features}}} & &tfidf     &   1.844 &  1.803 &   1.860 &    - &   \textbf{1.855} &  1.584 &   1.665 &    - &   \textbf{1.845} &  1.481 &   1.209 &    - \\
&&tfidf+lsa &   \textbf{2.026} &  1.870 &   1.992 &    - &   1.579 &  1.664 &   1.632 &    - &   1.727 &  1.602 &   1.606 &    - \\
\bottomrule
\end{tabular}
}
\caption[Aggregate topic coherence scores benchmark table]{Aggregate topic coherence ($\widehat{\text{PMI}}$) of unsupervised topic clusters (all clusterings and document features) at three resolution levels. Best clustering for each resolution level in boldface.}
\label{table:news:PMI_scores_10}
\end{table}

\vspace{-1cm}    
		    \paragraph{Comparison to external commercial classifications:}
		    \label{sec:news:external}
An alternative measure of topic cluster quality is to compare to other external classifications.
Here we use categorical labels of semantic information produced by two commercial text classification services: Google Cloud Platform's (GCP) Natural Language API, and Open Calais (OC) by Thomson Reuters. We compare our unsupervised topic clusters to the classes obtained by these commercial products (assumed to be trained on clean, human labelled data) using the Normalised Mutual Information (NMI) and the Adjusted Rand Index (ARI) as shown in Table~\ref{table:news:correspondence_scores}. 

\begin{table}[h]
\subfloat[Normalised Mutual Information]{
\centering
\scalebox{0.7}{
\begin{tabular}{clll|cccc|cccc|cccc}
\toprule
Commercial&  & & \textit{Resolution:} & \multicolumn{4}{c}{Fine} & \multicolumn{4}{c}{Medium} & \multicolumn{4}{c}{Coarse} \\
service&  & & \textit{Clustering:} &      MS &   Ward & k-means &    LDA &      MS &   Ward & k-means &    LDA &      MS &   Ward & k-means &    LDA \\
\midrule
GCP & & & baas &   0.356 &  0.335 &   0.345 &    - &   0.395 &  0.325 &   0.355 &    - &   0.400 &  0.339 &   0.345 &    - \\
&  & & d2v &   0.363 &  0.356 &   0.363 &    - &   0.391 &  0.351 &   0.380 &    - &   0.425 &  0.365 &   0.383 &    - \\
&  & & sbert &   0.347 &  0.312 &   0.320 &    - &   0.370 &  0.303 &   0.308 &    - &   0.381 &  0.286 &   0.258 &    - \\
&  & & tf &     - &    - &     - &  0.183 &     - &    - &     - &  0.161 &     - &    - &     - &  0.180 \\
&  \rot{\rlap{~\textit{Features}}}  & & tfidf &   0.353 &  0.339 &   0.369 &    - &   \textbf{0.404} &  0.332 &   0.377 &    - &   0.408 &  0.294 &   0.387 &    - \\
&  & & tfidf+lsa &   \textbf{0.389} &  0.350 &   0.370 &    - &   0.400 &  0.327 &   0.345 &    - &   \textbf{0.428} &  0.313 &   0.387 &    - \\
   \hline
OC & & & baas &   0.362 &  0.319 &   0.349 &    - &   0.392 &  0.315 &   0.350 &    - &   0.391 &  0.318 &   0.343 &    - \\
&  & & d2v &   0.372 &  0.349 &   0.363 &    - &   0.391 &  0.330 &   0.361 &    - &   \textbf{0.413} &  0.351 &   0.371 &    - \\
&  & & sbert &   0.354 &  0.309 &   0.321 &    - &   0.361 &  0.284 &   0.293 &    - &   0.394 &  0.259 &   0.225 &    - \\
&  & & tf &     - &    - &     - &  0.168 &     - &    - &     - &  0.175 &     - &    - &     - &  0.191 \\
&   \rot{\rlap{~\textit{Features}}} &  & tfidf &   0.364 &  0.341 &   0.354 &    - &   \textbf{0.410} &  0.310 &   0.371 &    - &   0.393 &  0.273 &   0.387 &    - \\
&  & & tfidf+lsa &   \textbf{0.389} &  0.341 &   0.372 &    - &   0.378 &  0.303 &   0.332 &    - &   0.411 &  0.296 &   0.344 &    - \\
\bottomrule
\end{tabular}
}
}

\subfloat[Adjusted Rand Index]{
\centering
\scalebox{0.7}{
\begin{tabular}{cl ll|cccc|cccc|cccc}
\toprule
 Commercial& & & \textit{Resolution:} & \multicolumn{4}{c}{Fine} & \multicolumn{4}{c}{Medium} & \multicolumn{4}{c}{Coarse} \\
 service&  & & \textit{Clustering:} &      MS &   Ward & k-means &    LDA &      MS &   Ward & k-means &    LDA &      MS &   Ward & k-means &    LDA \\
\midrule
GCP & & & baas &   0.171 &  0.152 &   0.163 &    - &   0.349 &  0.241 &   0.259 &    - &   0.483 &  0.295 &   0.306 &    - \\
&  & & d2v &   0.212 &  0.180 &   0.150 &    - &   0.360 &  0.252 &   0.254 &    - &   \textbf{0.559} &  0.494 &   0.415 &    - \\
&  & & sbert &   0.161 &  0.151 &   0.147 &    - &   0.319 &  0.292 &   0.227 &    - &   0.487 &  0.304 &   0.291 &    - \\
&  & & tf &     - &    - &     - &  0.127 &     - &    - &     - &  0.075 &     - &    - &     - & -0.004 \\
&   \rot{\rlap{~\textit{Features}}} &  & tfidf &   0.190 &  0.156 &   0.167 &    - &   \textbf{0.488} &  0.253 &   0.245 &    - &   0.520 &  0.195 &   0.312 &    - \\
&  &  & tfidf+lsa &   \textbf{0.301} &  0.173 &   0.167 &    - &   0.476 &  0.149 &   0.190 &    - &   0.552 &  0.151 &   0.342 &    - \\
   \hline
OC & & & baas &   0.171 &  0.150 &   0.162 &    - &   0.332 &  0.234 &   0.250 &    - &   0.405 &  0.251 &   0.282 &    - \\
&  & & d2v &   0.215 &  0.176 &   0.157 &    - &   0.336 &  0.229 &   0.224 &    - &   \textbf{0.468} &  0.428 &   0.408 &    - \\
&  & & sbert &   0.173 &  0.156 &   0.159 &    - &   0.311 &  0.252 &   0.213 &    - &   0.462 &  0.269 &   0.253 &    - \\
&  & & tf &     - &    - &     - &  0.102 &     - &    - &     - &  0.114 &     - &    - &     - &  0.019 \\
&  \rot{\rlap{~\textit{Features}}} &  & tfidf &   0.197 &  0.149 &   0.147 &    - &   \textbf{0.422} &  0.194 &   0.224 &    - &   0.409 &  0.164 &   0.301 &    - \\
&  & & tfidf+lsa &   \textbf{0.284} &  0.161 &   0.160 &    - &   0.399 &  0.146 &   0.171 &    - &   0.457 &  0.171 &   0.269 &    - \\
\bottomrule
\end{tabular}
}
}
\caption[Correspondence scores against commercial classification service labels]{Unsupervised topic clusters (all features and clusterings) at three resolution levels scored against two commercial classification labels: Google Cloud Platform (GCP) and Open Calais (OC). Best clustering for each resolution level in boldface.} 
\label{table:news:correspondence_scores}
\end{table} 

From the NMI and ARI scores for all our experiments, we find that MS graph-partitioning provides the best correspondence to the external classifications followed by k-means and Ward. LDA provides the worst scores against the external labels. 
Again, there is no clear winner among the features: \textit{tfidf+lsa}, \textit{tfidf} and \textit{d2v} (all with MS) produce the best results against external classes at the F, M and C levels.
Among the Bert variants, {\it baas} performs better than {\it sbert} but they do not outperform {\it d2v}.

\begin{figure}[h]
	\centering
\includegraphics[width=\linewidth]{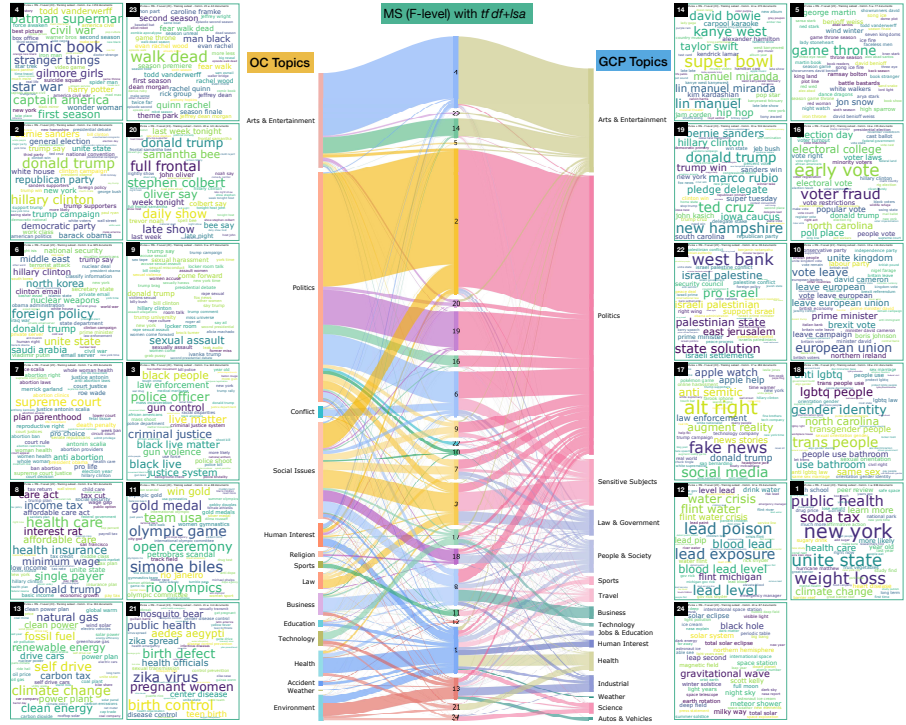}
\caption[Sankey diagram with wordclouds for MS of $\mathcal{G}_\textrm{\small \it tfidf+lsa}$]
{Sankey diagram and wordclouds for MS partitions of $\mathcal{G}_\textrm{\small \it tfidf+lsa}$ and the external labels of the commercial services GCP and OC.}
\label{fig:news:OC_GCP_tidf}
	\end{figure}

As visualisation aids, we use multilevel Sankey diagrams to represent relationships between partitions, and wordclouds to represent the content of topic clusters. In Figure~\ref{fig:news:OC_GCP_tidf}, we show the mapping of the external classes of the commercial services OC and GCP against the best unsupervised clustering at the F level, given by MS of $\mathcal{G}_\textrm{\small \it tfidf+lsa}$. Our clustering shows strong agreement with the large categories in both OC and GCP with additional detail in several groupings.

Overall, our results show that MS usually performs better than the other clustering algorithms both in terms of intrinsic word consistency and when comparing to commercial hand labelled external classes.

\vspace{-.5cm}  
    	\subsection{Qualitative analysis of the topics}
    	\label{sec:news:qualitative}

	\begin{figure}[h]
	\centering
\includegraphics[trim={5cm 0.10cm 5cm 0.2cm}, clip, width=\linewidth]{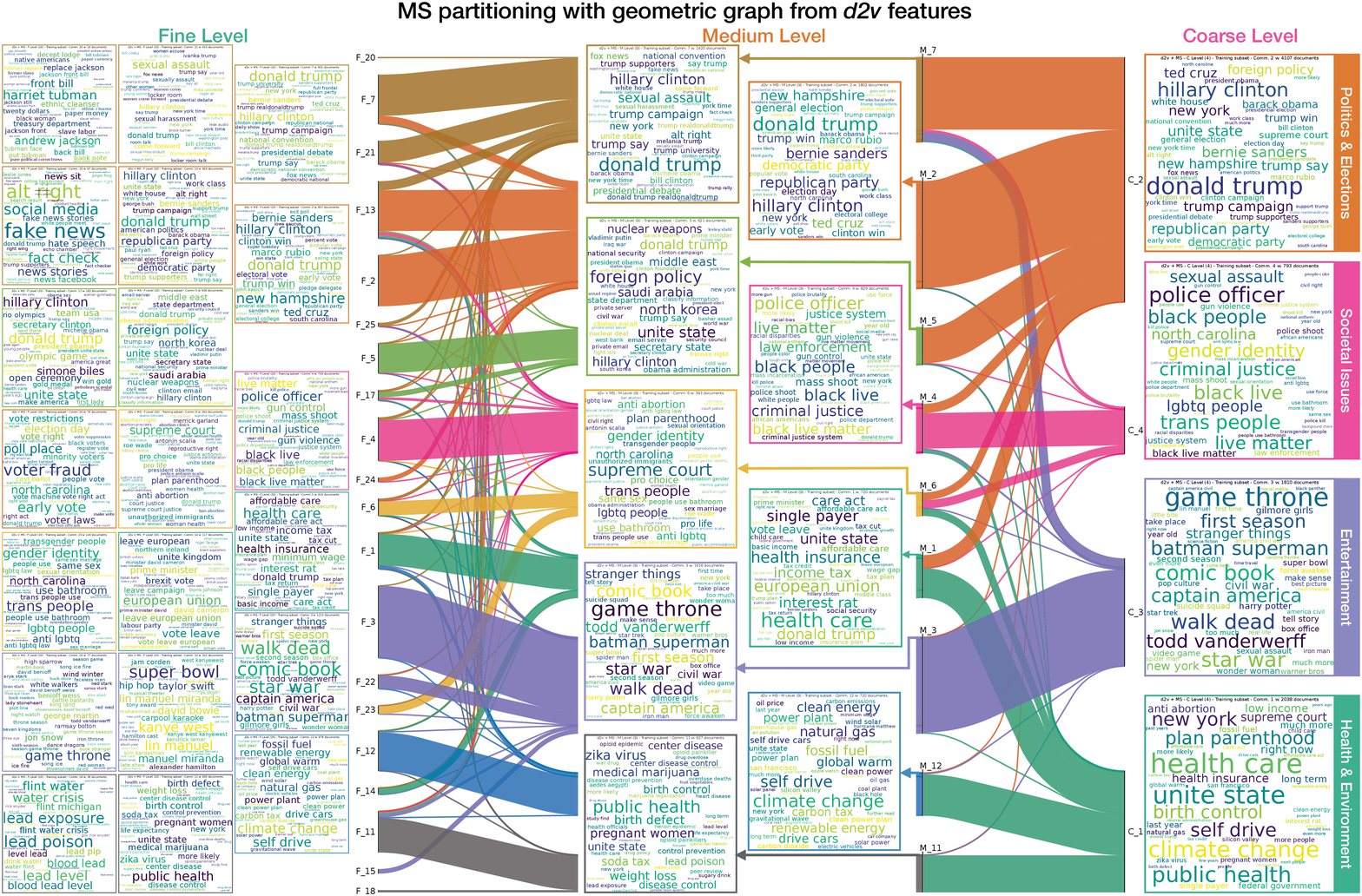}
\caption[Multilevel Sankey diagram with wordclouds for MS analysis of $\mathcal{G}_\textrm{\small \it d2v}$]
{Multilevel Sankey diagram and wordclouds of the MS partitions of $\mathcal{G}_\textrm{\small \it d2v}$ at the three resolution levels (fine, medium, coarse).}
\label{fig:news:d2v_ms_sankey}
	\end{figure}

Using a multilevel Sankey diagramme and wordclouds, Figure~\ref{fig:news:d2v_ms_sankey} shows the MS topic clusters obtained from $\mathcal{G}_\textrm{\small \it d2v}$  at the three resolutions: fine, medium, coarse. We find four main topics at the C level, which, using information from the wordclouds, we label as `Politics~\&~Elections', `Societal Issues', `Entertainment', and `Health~\&~Environment'.
The last one divides into three communities on the medium resolution: `Public health~\&~Medicine', `Energy~\&~Environment', and `Healthcare~\&~Insurance'. Among those, the last one also involves some contribution from the `Politics~\&~Elections' community over the discourse of healthcare policy of the presidential candidates. 

A similar mixed contribution is observed on the M4, a finer resolution community of `Societal Issues'. M4 contains news articles related to `Black Lives Matter (BLM)' and lies close to `Politics~\&~Elections'. 
The other finer level of `Societal Issues' is M6 that mixes `Politics~\&~Elections' and `Health~\&~Environment'  as it involves `gender identity' issues in the US as well as discussions around `planned parenthood'.  This topic is a good example of the advantage of having intrinsically quasi-hierarchical partitions instead of forced hierarchies as in Ward's hierarchical clustering. 
Indeed, in a strictly hierarchical clustering, we would have either lost M6 completely engulfed by another topic, or we would have obtained less coherent topic cluster(s) on other levels.
The flexibility of quasi-hierarchical structures, however, allows us to extract topics independently to the other resolutions.

Figure~\ref{fig:news:d2v_ms_sankey} also shows that the top community in the M level (M7) is largely conformed by`Politics~\&~Elections' with small involvement of `Entertainment'. Since this topic is related to the election campaigns, the involvement of `Entertainment' reflects the role of media in campaigns through e.g., political speeches, interviews and debates in TV shows as well as allegations of`fake news' by one of the candidates who targeted the media industry (a small finer community F25).
Although we only describe a few examples of interesting relationships across the three levels of resolution, all communities have distinct and high quality topics as reflected in their wordclouds.

    \begin{figure}[h]
    \centering
\includegraphics[trim={8.5cm 0.10cm 8cm 0.35cm}, clip, width=\linewidth]{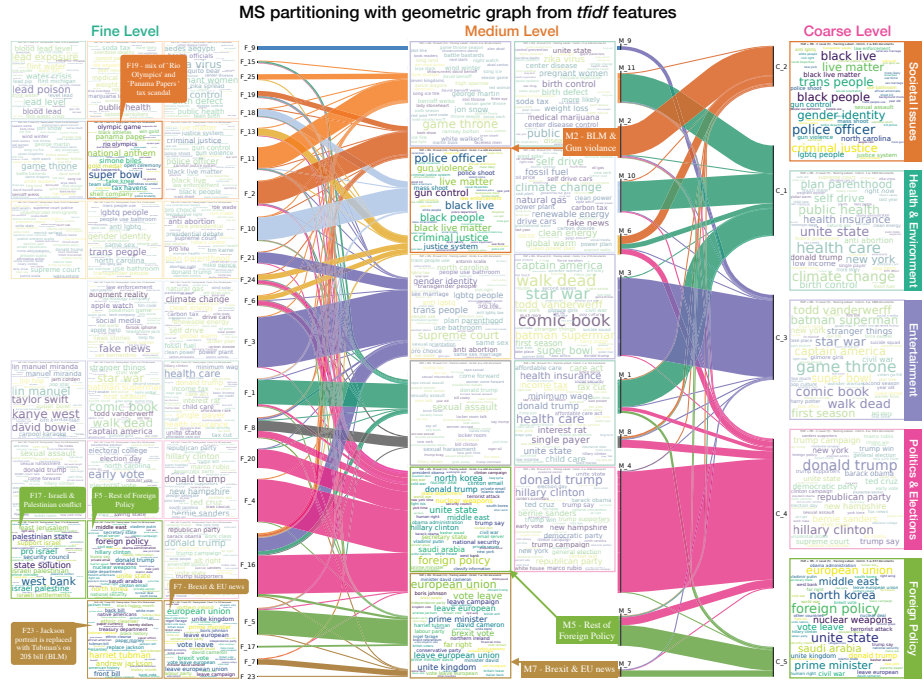}
\caption{Multilevel Sankey diagram and wordclouds for the three MS partitions of $\mathcal{G}_\textrm{\small \it tfidf}$. The highlighted wordclouds correspond to topic clusters mentioned in the main text.}
\label{fig:news:tfidf_ms_sankey}
	\end{figure}

To illustrate the importance of the choice of features for text embedding,
in Figure~\ref{fig:news:tfidf_ms_sankey} we examine the MS topic clusters obtained from $\mathcal{G}_\textrm{\small \it tfidf}$, i.e., from BoW-based features.
Four of the five communities at the coarsest level of MS with \textit{tfidf} are consistent with the four coarse level topics obtained by MS with \textit{d2v}.
The additional topic is `Foreign Policy'.
Interestingly, a very similar cluster appears as a distinct grouping in the M level in the \textit{d2v} partitions (M5 in the Figure~\ref{fig:news:d2v_ms_sankey}) under `Politics~\&~Elections'.
The `Foreign Policy' cluster in Figure~\ref{fig:news:tfidf_ms_sankey} includes as a subcluster
`Brexit and EU' news, one of the most unique and important events in 2016, and a subcluster that divides into a tiny but distinct topic of `Israeli and Palestinian conflict', thus signalling the consistency of topic clusters.
On the other hand, within the `Brexit and EU' we find a finer subcluster (F23) with news articles mentioning that Andrew Jackson's portrait is being replaced by Harriet Tubman's on \$20 bills, thus indicatig that the `BLM' movement has been mistakenly grouped under M7.
Hence BoW clusters at fine levels can lead to mixed topics due to coincidental use of specific words and the reliance on pure word counts, instead of contextual semantic predictions. A good example for this problem is F19, which mixes three unrelated topics of news articles from events like `Panama papers', `super bowl', and `Rio Olympics' which share a list of common word tokens.

For further inspection of the resulting topic clusters, we provide a time line of monthly clusters accessible though interactive plots on \url{http://bit.ly/Vox2016}.

	\section{Conclusion}
	\label{sec:news:conclusion}

In this paper, we have presented a graph-based methodology for unsupervised topic extraction from text, which uses text vector embeddings and multiscale graph partitioning.   Using a corpus of 9,000 US news articles collected during 2016, we have compared our graph-based clustering to other widely used graph-less clustering methods, such as k-means, hierarchical clustering and LDA topic modelling. Taking advantage of the recent significant improvements in natural language processing, we have evaluated different document embedding vectors, from the simplest BoW features to Doc2Vec to the recent Bert based alternatives. We benchmarked our results using measures of intrinsic word consistency, as well as comparisons to external commercial classifications in order to quantify the quality of the resulting topic clusters. 

Using our quantitative analysis, we concluded that MS partitioning outperforms k-means, Ward and LDA clusters in almost all cases. We also observed that {\it d2v} embeddings are still among the best methods, outperforming the Bert variants on our data set. Most surprisingly, the traditional BoW based {\it tfidf} and {\it tfidf+lsa} features are also successful, although 
their performance is less robust than {\it d2v} with some impure topic clusters dominated by strong word counts.

The qualitative analysis of the topic clusters displays quasi-hierarchical consistency in the topics, allowing for flexible subtopics to emerge at finer resolutions. Our analysis also shows
that the cluster content is affected by both the features and the partitioning methods, especially at the finer levels.   
A conclusion of our qualitative analysis is that quantitative benchmarks based on word consistency or comparisons to external commercial classifications do not capture fully the quality of the topics. Future work should be aimed at improving the quantitation of content quality. For instance,  the aggregation of topic coherence per topic does not take repetition into account, hence it does not penalise a partition with multiple topic clusters with similar content (see e.g.,  
`Politics~\&~Elections' and `Foreign Policy' in the same partition in Figure~\ref{fig:news:tfidf_ms_sankey}).

Overall, we conclude that our graph partitioning approach using Markov Stability works well for topic clustering by using community detection on geometric graphs obtained from high-dimensional embedding vectors of documents. Its advantages include the possibility of obtaining topic clusters at different levels of granularity,  with a flexible quasi-hierarchy of related topics and robust results. 
We also conclude that {\it d2v} features serve well for the objective of topic clustering with good quantitative scores and low topic confusion in our  qualitative analysis.

\bibliographystyle{splncs03}
\bibliography{zotero_abbrv}

\end{document}